\documentclass[pdflatex,sn-mathphys-num]{sn-jnl}% Math and Physical Sciences Numbered Reference Style 
\usepackage{graphicx}%
\usepackage{multirow}%
\usepackage{amsmath,amssymb,amsfonts}%
\usepackage{amsthm}%
\usepackage{mathrsfs}%
\usepackage[title]{appendix}%
\usepackage{xcolor}%
\usepackage[sort&compress]{natbib}
\usepackage{textcomp}%
\usepackage{manyfoot}%
\usepackage{booktabs}%
\usepackage{algorithm}%
\usepackage{algorithmicx}%
\usepackage{algpseudocode}%
\usepackage{listings}%
\usepackage[strings]{underscore}
\usepackage{bbm}
\usepackage{xspace}
\usepackage{float}
\usepackage[inline]{enumitem}
\usepackage{mathtools}
\usepackage{booktabs}
\usepackage{subfig}
\usepackage{siunitx}
\usepackage{courier}
\usepackage{gensymb}
\makeatletter
\DeclareRobustCommand\onedot{\futurelet\@let@token\@onedot}
\def\@onedot{\ifx\@let@token.\else.\null\fi\xspace}
 
\def\ie{\emph{i.e}\onedot} 
 
\def\etc{\emph{etc}\onedot}

\newlist{inlist}{enumerate*}{1}
\setlist[inlist]{label=(\arabic*)}
\newlist{inlistalpha}{enumerate*}{1}
\setlist[inlistalpha]{label=(\alph*)}
\newcolumntype{L}[1]{>{\raggedright\arraybackslash}p{#1}}
\newcolumntype{R}[1]{>{\raggedleft\arraybackslash}p{#1}}
\newcolumntype{C}[1]{>{\centering\arraybackslash}p{#1}}
\newcommand{\R}{\ensuremath{\mathbb{R}}}

\definecolor{chcolor}{RGB}{46, 179, 81}

\theoremstyle{thmstyleone}%
\theoremstyle{thmstyletwo}%

\theoremstyle{thmstylethree}%

\begin{document}

% \title[]{Take Any Shot! Controlling Robotic X-ray Systems using a Language-aligned Segment Anything Model}
% \title[]{A Language-promptable Voice Interface for Controlling Robotic X-ray Systems in Surgery}
\title[]{Intelligent Control of Robotic X-ray Devices using a Language-promptable Digital Twin}
% =======
% \title[]{Commanding Robotic X-ray Systems using a Language-prompted Segment-anything Digital Twin}
% \title[]{Foundation Models Unify Full-body Voice User Interfaces for Robotic X-ray Systems}

% Language Interfaces for Robotic X-ray Imaging Through Interactive Segment Anything

\author*[1]{\fnm{Benjamin~D.} \sur{Killeen}}\email{killeen@jhu.edu}
\author[1]{\fnm{Anushri} \sur{Suresh}}\email{asures13@jhu.edu}
\author[1]{\fnm{Catalina} \sur{Gomez}}{\email{cgomezc1@jhu.edu}}
\author[1]{\fnm{Blanca} \sur{\'I\~nigo}}{\email{binigo2@jhu.edu}
\author[2]{\fnm{Christopher} \sur{Bailey}}\email{Christopher.Bailey@jhmi.edu}
\author[1]{\fnm{Mathias} \sur{Unberath}}\email{unberath@jhu.edu}

%Times New RomanTimes New Roman Remember to mark corresponding author in final submission.
\affil*[1]{\orgdiv{Laboratory for Computational Sensing and Robotics}, \orgname{Johns Hopkins University}, \orgaddress{\city{Baltimore}, \postcode{21218}, \state{MD}, \country{USA}}}
% \affil[1]{\orgname{Johns Hopkins University}, \orgaddress{\city{Baltimore}, \postcode{21210}, \state{MD}, \country{USA}}}
\affil[2]{\orgdiv{Department of Interventional Radiology}, \orgname{Johns Hopkins University}, \orgaddress{\city{Baltimore}, \postcode{212187}, \state{MD}, \country{USA}}}
% \affil[2]{\orgdiv{Department of Orthopaedic Surgery},}

\abstract{\textbf{Purpose:}
Natural language offers a convenient, flexible interface for controlling robotic C-arm X-ray systems, making advanced functionality and controls easily accessible. 
However, enabling language interfaces requires specialized AI models that interpret X-ray images to create a semantic representation for language-based reasoning. The fixed outputs of such AI models fundamentally limits the functionality of language controls that users may access.
Incorporating flexible and language-aligned AI models that can be prompted through language control facilitates more flexible interfaces for a much wider variety of tasks and procedures.

\textbf{Methods:}
Using a language-aligned foundation model for X-ray image segmentation, our system continually updates a patient digital twin based on sparse reconstructions of desired anatomical structures. This allows for multiple autonomous capabilities, including visualization, patient-specific viewfinding, and automatic collimation from novel viewpoints, enabling complex language control commands like ``Focus in on the lower lumbar vertebrae.''

% We present an English language voice interface for controlling robotic X-ray devices that supports automatic localization, collimation, and visualization of arbitrary anatomical structures, including from novel viewpoints. Our fully-integrated system uses FluoroSAM, a language-prompted segment-anything model for X-ray images, to analyze current and past images based on user input, enabling commands like, ``Show me the lower lumbar vertebrae'' or ``Move over the right femur.''

\textbf{Results:} In a cadaver study, multiple users were able to visualize, localize, and collimate around structures from across the torso region using only verbal commands to control a robotic X-ray system, with 84\% end-to-end success.
In \emph{post hoc} analysis of randomly oriented images, our patient digital twin was able to localize 35 commonly requested structures from a given image to within $51.68 \pm 30.84$\,mm, which enables localization and isolation of the object from arbitrary orientations.

% , approximately 200\,mm at the imaging center.
\textbf{Conclusion:} Overall, we show how intelligent robotic X-ray systems can incorporate physicians' expressed intent directly. Existing foundation models for intra-operative X-ray image analysis exhibit certain failure modes. Nevertheless, our results suggest that as these models become more capable, they can facilitate highly flexible, intelligent robotic C-arms.
}




\maketitle

\section{Introduction}
\label{sec:introduction}

% So far AI capabilities are being developed for X-ray guided surgery that are highly specialized. Natural language provides a way to make these high-level capabilities ready-to-hand, but incorporating myriad specialized models into a single natural language interface is not viable. Capabilities driven by foundation models, which are agnostic to the specific task, can be incorporated into a natural language interface to provide a more flexible and general-purpose interface for controlling robotic X-ray systems in surgery.

Natural language offers an appealing interface for commanding robotic X-ray devices in surgery, allowing physicians to express their imaging needs rather than execute them manually~\cite{killeen2024take}.
When equipped with artificial intelligence (AI) models capable of analyzing intra-operative images, voice-controlled robotic C-arms effectively become intelligent assistants for image-guided surgery, with the potential to reduce radiation exposure~\cite{kausch2020toward}, avoid complications~\cite{killeen2023autonomous}, streamline procedures~\cite{killeen2023pelphix}, and improve overall patient outcomes.
So far, however, intelligent systems for X-ray image-guided surgery have relied on specialized, task-specific models with fixed outputs, limiting their general application.
Meanwhile, voice interfaces for general robotics are rapidly accelerating due to the rise of multi-modal foundation models~\cite{kawaharazuka2024real}, which are characterized by large-scale training and generalizability for a wide range of downstream tasks~\cite{bommasani2021on}.
Although foundation models have been developed for the X-ray domain~\cite{chen2401chexagent, ma2024segment, killeen2024fluorosam}, they are generally limited to diagnostic chest images, and it remains unclear how to incorporate them into a voice-user interface for commanding robotic C-arms more generally.

Here, we leverage an X-ray foundation model to support intelligent capabilities in a voice-controlled robotic C-arm, including visualization, collimation, and patient-specific viewfinding. Our approach continually updates a patient digital twin using images acquired during surgery. A large language model (LLM) interprets spoken commands, like ``focus on the lower lumbar vertebrae'' and extracts the intended action. Low-level actions, like ``roll over 30 degrees'' are converted directly to joint movements by the LLM, while high-level actions are associated with a language prompt, \ie ``lower lumbar vertebrae,'' potentially based on past commands. The LLM then sends the action and prompt to the digital twin, which uses a multi-modal segment-anything model (FluoroSAM \cite{killeen2024fluorosam}) to analyze past images and aggregate 3D information about the desired anatomy. This enables future acquisitions from unseen viewpoints to have appropriate collimation for the desired anatomy, limiting radiation exposure while still ensuring the structure remains in view. This also allows for automatic viewfinding, where the robotic C-arm can adjust its position and orientation toward a desired anatomy, based on the current understanding.

We demonstrate our fully integrated system in a cadaver study with spoken prompts and in post-hoc analysis of randomly acquired X-ray images registered to a full-body CT. Our results demonstrate the robustness of the voice interface in real-time, with a success rate of 83.54\%, and the accuracy of the digital twin in aligning with ground truth structures in the CT. In \emph{post hoc} analysis of 1990 possible image subsets, the digital twin yields an average $51.68 \pm 30.84$\,mm localization error, with a 3D bounding box precision and recall of 0.26 and 0.70 for tested prompts, respectively. Our work paves the way for general-purpose AI surgical assistants in image-guided surgery, supporting a wide range of high-level capabilities by incorporating.

\begin{figure}[t]
  \centering
  \includegraphics[width=\linewidth]{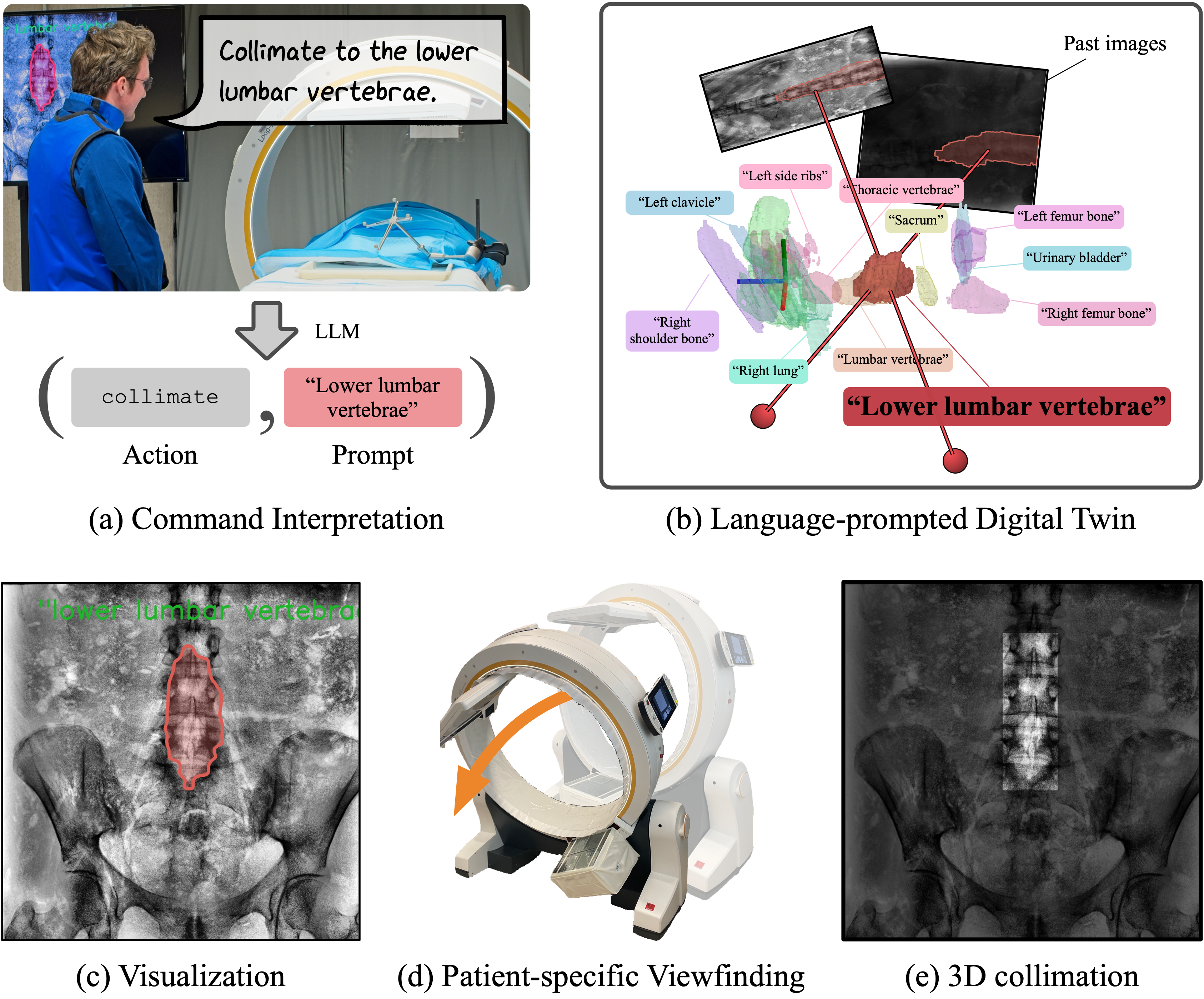}
  \caption{We present a natural language interface for commanding robotic X-ray devices using a multi-modal foundation model for X-ray imaging. Our approach, which we demonstrate in a real-time cadaver study (a), uses a large language model to parse the desired action and suitable prompt from the spoken input. The digital twin (b) uses FluoroSAM~\cite{killeen2024fluorosam} to segment anatomies based on the prompt, supporting real-time visualization (c), patient-specific viewfinding (d), and 3D collimation.
  }
  \label{fig:overview}
\end{figure}

\section{Related Work}
\label{sec:related-work}
% Discussion of foundation models facilitating voice interfaces for general robots, then also voice or text interfaces in the medical context using foundation models (e.g. chexpert, etc)
% \cite{kim2021development} discusses VUIs
Previous works have explored integrating natural language into robotic systems, given the promise of more intuitive and flexible interactions with users, particularly through voice and text commands~\cite{tellex2020robots}. Such systems may demand real-time language understanding, interpretation, and generation for inference while operating and executing commands constrained by real-world physics and the environmental context~\cite{vemprala2024chatgpt}. Recent developments in foundation models are appealing for robotic applications considering the wide range of downstream tasks they can be adapted to and their ability to process multimodal data~\cite{kawaharazuka2024real}. The combination of existing language models with vision-based models enables the development of robots that can engage in conversations with users through textual interfaces~\cite{dai2024think} or real-time instructions~\cite{lynch2023interactive} for object navigation,  modifying robot trajectories~\cite{martinez2024hey,bucker2022reshaping}, and fine-grained manipulation~\cite{shridhar2022cliport}.  
% efforts in building datasets to adapt general FM
% RGB data as the visual input - more similar to the datasets used to train general FM 
Importantly, these language-conditioned models ground user text queries to the robot's visual observations of its environment, generating actions within the robot's capabilities.

Likewise, foundation models in medicine can interpret multimodal data and offer communication via multiple modalities (text, visualizations, etc.), enabling novel human-machine interactions and improved generalization abilities~\cite{moor2023foundation}. The development of medical foundation models has been particularly successful in radiology, especially for chest X-rays, where abundant image-text paired data facilitates fine-tuning of general-domain models. These models can leverage existing image-based analysis methods~\cite{gao2023synthetic} and clinical large language models (e.g., BioGPT~\cite{luo2022biogpt}) to train vision-language assistants with conversational abilities~\cite{li2024llava} and successful at both image interpretation and textual understanding~\cite{chen2401chexagent}, among a wide range of tasks identified through collaborative work with medical experts~\cite{yildirim2024multimodal}. 
Segmentation is another fundamental task in medical image analysis where foundation models have shown improvements with respect to specialized models. Existing medical Segment Anything models~\cite{ma2024segment,shen2024fastsam3d} follow the fine-tuning strategy of a general-domain foundation model for segmentation on image-mask pairs covering multiple imaging modalities and disease types. While promptable segmentation models can adjust to different use cases and users' needs, point-based prompts can be ambiguous and bounding boxes require expert input. Instead, text prompts enable broader accessibility, especially in scenarios like surgery where voice commands (mapped to text) can streamline operating room (OR) workflows~\cite{kim2021development}. 
FluoroSAM~\cite{killeen2024fluorosam} supports text-only prompting for segmentation of anatomical structures in X-rays, for which scalable data generation in simulation was needed~\cite{killeen2023insilico, killeen2023pelphix, unberath2018deepdrr}.

% more works on diagnostics: 
% endoscopy data - RGB images, more likely to work.  

% how to include/leverage the flexibily of multimodal FM into an interface for interventional robots is unclear/ in interventional radiology settings given the constraints of navigation. 
% idea of leveraging speech motivated by: 
% \cite{kim2021development}: a voice-controlled virtual smart assistant in the OR
Despite multiple efforts leveraging the appealing features and flexibility of multimodal foundation models, it remains unclear how these benefits translate beyond diagnostic tasks to interventional procedures. In particular, how to include language-based interactions in interventional radiology is crucial for developing smart OR assistants grounded on the complexities of X-ray image-guided procedures.

% \todo{Make sure to cite: \cite{killeen2024stand} (Can discuss data collection of human interactions) \cite{gao2023synthetic, unberath2018deepdrr} for X-ray analysis, \cite{killeen2023pelphix} for adding contextual awareness to intelligent C-arms, \cite{killeen2023insilico} for learning from simulation, \cite{ding2024digital}, maybe \cite{zhang2024straighttrack} but it's a stretch.}

\section{Methods}
\label{sec:methods}

Our approach consists of a command interpretation system and a digital twin that supports multiple high-level capabilities using a language-aligned foundation model. Following~\cite{killeen2024take}, we define a communication protocol for an LLM to interpret natural language as machine-readable actions. The user gives natural commands like ``Show me the right lung,'' and the LLM returns a machine-readable action, in this case a \texttt{highlight} action with the text-prompt ``right lung.'' In our experiments, we used a lavalier microphone clipped to the user's lead apron to record speech, with a mute button to prevent cross-talk. In practice, we envision the microphone being unmuted using a foot pedal or other sterility-preserving mechanism. Live speech-to-text is accomplished via OpenAI Whisper. Using low-level commands, such as ``roll over 30 degrees'' the user can adjust independent axes and take an X-ray, as described in~\cite{killeen2024take}.
High-level commands trigger one of three new high-level actions using language-aligned foundation model: visualization (\texttt{highlight}), collimation (\texttt{collimate}), and patient-specific viewfinding (\texttt{view}).
Visualization is a straightforward function for displaying, in real time, the segmentation of a given anatomy on the current image. Collimation and patient-specific viewfinding rely on the digital twin to localize the desired structure in 3D.
For example, the portion of the instruction that specifies the \texttt{view} action reads:
\begin{lstlisting}
A `view` action requests a specific view:

action;view;view_name;prompt

where `view_name={ap|lateral|current}`
and `prompt` is derived from the user input to be used as a language prompt for a segmentation model. `prompt` should be a concise description of the desired anatomy or structure. If no anatomy is specified, use `prompt=current`.
\end{lstlisting}
%which are 1,280 words (3,706 tokens)

In total, our instruction set and examples are 1655 words (4356 tokens) total. This is supported by the maximum input token length of the GPT-4o (128,000) model used in our experiments, and results in an LLM latency of less than 1 second. This is small compared to the robotic movement and acquisition times of the Loop-X, which are typically on the order of 10 or more seconds. For safety reasons, the Loop-X requires users to confirm these actions using a physical button; however, during our experiments, the user never initiated actions directly from the Brainlab user interface. This physical confirmation is a specific requirement for the Loop-X device, but future systems might remove confirmation for non-radiation actions. % as they build greater trust with the surgeon~\tocite{}.

\subsection{A Digital Twin Based on a Language-aligned Foundation Model}
\label{sec:digital-twin}

We describe a highly flexible patient digital twin that incorporates casually acquired images to reconstruct desired anatomical structures in 3D. This is accomplished through text-prompted segmentation of past-structures using FluoroSAM~\cite{killeen2024fluorosam}, a segment-anything model for X-ray imaging that uses language to disambiguate overlapping structures. Traditionally, segment-anything models (SAMs) predict a valid mask for a given prompt, which may be a point, mask, or 2D bounding box. FluoroSAM incorporates a CLIP~\cite{radford2021learning, wang2022medclip} embedding of an anatomy description. We examine the use of FluoroSAM without additional point prompts, using only the extracted prompt from the LLM to obtain a segmentation.

Formally, let $f(\mathbf{u},I,t)$ denote the value of the FluoroSAM logits at $\mathbf{u}\in \R^2$ for image $I$ with prompt $t$. Let $I_{0} \in \R^{H_0 \times W_0}$ and $\mathbf{P}_0 = \mathbf{K} [\mathbf{R}_0 \vert \mathbf{t}_0]$ denote the image being examing as well as its projection matrix relative to a fixed coordinate system, such as the optical marker in Fig.~\ref{fig:overview}. Following \cite{killeen2023autonomous}, to support 3D reconstruction we identify a set of images that have a minimum acute angle of 30\degree. If two images are available from similar viewpoints ($<10^{\circ}$ angle), we take the most recent, so as to maintain an up-to-date digital twin. Practically, this is reasonable given the routine workflows in many image-guided procedures, which alternate between views such as the anteroposterior (AP) and lateral~\cite{killeen2023pelphix, opfermann2021feasibility}. If such images are not available, such as at the start of the procedure, our system still supports visualization, collimation, and translation of the imaging center based on $f(I_0,t)$. Let $I_0, I_1,\dots,I_{n-1}$ denote the images identified from the history, including the current image. $n$ is typically between 2 and 5, given the constraints on image selection. We define the sparse reconstruction of the anatomy as the set of 3D points where the mean value of $f$ is at least 0.5, and the point is present in at least 2 masks:
\newcommand{\projx}[1]{\mathbf{P}_{#1}\tilde{\mathbf{x}}}
\begin{equation}
  \label{eq:reconstruction}
  \textbf{X} = \left\{\mathbf{x} \in \mathbb{R}^3 \;\middle|\; 
  \vec{0} \leq \projx{0} < \begin{bmatrix}
      W_0\\
      H_0
      \end{bmatrix} \text{ and }  \left|\mathcal{I}_{\mathbf{x}}\right| \geq 2 \text{ and } \frac{1}{\left|\mathcal{I}_{\mathbf{x}}\right|} \sum_{i \in \mathcal{I}_{\mathbf{x}}} f\left(\projx{i}, I_i, t\right)\right\}
\end{equation}
where $\mathcal{I}_{\mathbf{x}} = \{i \mid f(\projx{i}, I_i, t) > 0.5\}$ is the set of images for which a point $\mathbf{x}$ is in the mask. This is computed on GPU by backprojecting each heatmap value through a patient volume with 3\,mm isocentric spacing, which is sufficient given the 0.3\,mm detector pixel size on our device. Fig.~\ref{fig:overview}b shows sparse reconstructions of several prompts over the torso, including ``lower lumbar vertebrae'' from the given images.
As can be observed, the quality of these reconstructions depends in large part on the quality and consistency of the segmentation mask across images, but we show in our experiments that they are sufficient to locate the centroids of observed structures, and they enable significant collimation of desired structures. Our system supports viewfinding based on AP and lateral orientations, as well as previous viewpoints.

\section{Experiments}
\label{sec:experiments}

We evaluate our approach in terms of three aspects, based on a cadaveric imaging study using a Brainlab Loop-X robotic X-ray device. First, we examine the 2D performance of FluoroSAM, which was trained using digitally reconstructed radiographs, on real images. Since the current version of FluoroSAM was designed with point-based prompts in mind, it has notable failure modes when using text-only prompting. Second, we evaluate the accuracy of our 3D digital twin for prompts where FluoroSAM achieves a sufficient DICE score. Finally, we examine the real-world usability of the fully integrated system for spoken prompts across a variety of anatomies and views.
The cadaveric specimen included the torso section from the mid-femur to the T2 vertebra, not including the arms, from a 60 year-old female donor with a living height of 157\,cm and living weight of 50\,kg. The specimen was thawed at 4\degree C for 6 days prior to the study. All fluoroscopic images were acquired with navigation relative to a fixed patient array. To obtain complete ground truth masks, we stitched together 4 navigated cone-beam CT images acquired with the Loop-X immediately following the study and project organ segmentations~\cite{wasserthal2023totalsegmentator} onto each image. These are combined according to the specifications of each prompt, such as by joining L3, L4, and L5 for the ``lower lumbar vertebrae.'' 80 X-ray images were acquired over the course of the study, of which 46 were from unique viewpoints. The specimen was not moved throughout the study, aside from slight soft tissue deformation due to settling.

\subsection{FluoroSAM Performance}
\label{sec:eval-fluor-perf}

We evaluate FluoroSAM's performance using prompts obtained from an attending interventional radiologist that describe primary structures in the torso and pelvic region. Overall, FluoroSAM is able to segment large structures based on text-only prompting, but it fails to localize small, repeating structures and small organs. This is reasonable, since small organs are generally not visible without contrast agent. Moreover, a single image may not provide enough visual context to distinguish between similar vertebrae levels (0.04 DICE) or rib bones (0.02 DICE) without additional point prompting, as in \cite{killeen2024fluorosam}. However, we do observe reasonable performance when segmenting larger structures, such the vertebrae (0.78 $\pm$ 0.15), vertebrae sections (0.72 $\pm$ 0.18 lumbar and $0.64 \pm 0.20$ thoracic), the ribcage ($0.54 \pm 0.12$), \etc. Fig.~\ref{fig:fluorosam-results} shows the DICE score for the top 40 prompts, which we use in our study, and example predictions. These include organs which are in the base TotalSegmentator~\cite{wasserthal2023totalsegmentator} classes from which FluoroSAM was trained, as well as groups of organs never seen before, like ``lower lumbar vertebrae,'' which we define as L3 - L5. Table~\ref{tab:fluorosam-results} details the DICE score and centroid error for the top 40 prompts used. We also note that FluoroSAM is able to localize many structures, with an average centroid error of $58.87 \pm 50.29$\,mm among the top 40 DICE prompts. The detector size is 430\,mm square.

\begin{figure}[t]
  \centering
  % \subfloat[FluoroSAM Performance]{
  %   \label{fig:fluorosam-dice}
  %   \includegraphics[height=3in]{images/fluorosam-dice.png}
  % }
  % \subfloat[Example Prompts]{
  %   \label{fig:fluorosam-examples}
  %   \includegraphics[height=3in]{images/examples.png}
  % }
  \includegraphics[width=\linewidth]{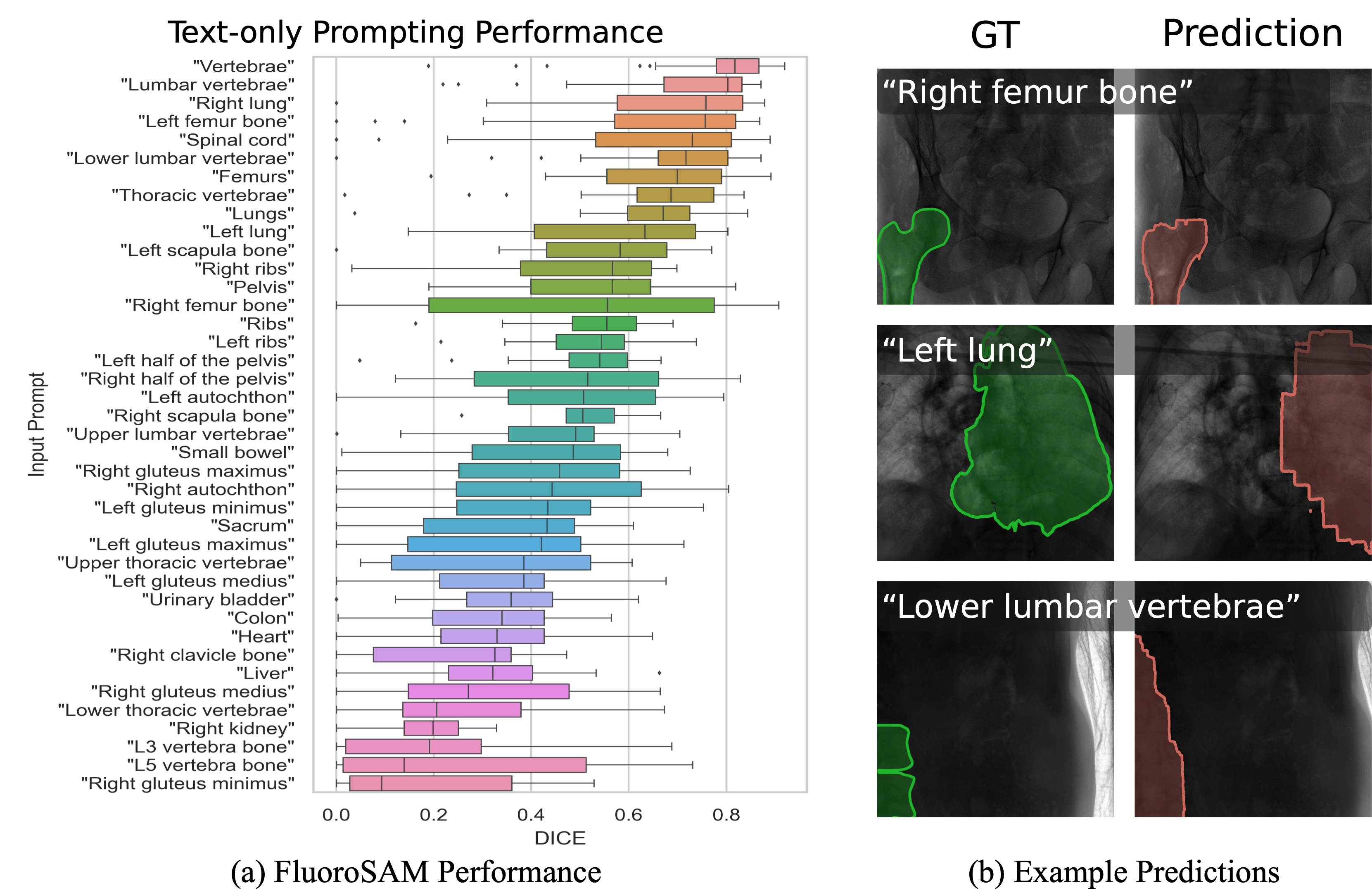}
  \caption{Performance and example masks for text-only prompting with FluoroSAM. Although certain classes struggle without additional point-based prompts, the model correctly localizes many structures based on CLIP~\cite{radford2021learning, wang2022medclip} embeddings of natural language prompts, extracted by the LLM protocol, including unseen prompts like ``lower lumbar vertebrae.'' }
  \label{fig:fluorosam-results}
\end{figure}

\subsection{Digital Twin Reconstruction}

We evaluate our digital twin in terms of its ability to localize and isolate desired structures in a given image, using randomly selected images as secondary shots. Using each of the 46 unique images in our study as the primary image, we randomly sample additional images from $n = 2$ to $n = 5$, totaling 1990 unique subsets of images with common structures for which FluoroSAM's DICE score exceeds 0.3, to isolate the ability of the digital twin to localize structures given a reasonable segmentation. We ignore the acquisition time for the purpose of evaluation. We evaluate the centroid error of the desired anatomy relative to its overlap with the ``current'' image, in alignment with physicians' expectations. For these anatomy, our system is able to localize desired structures in 3D to within $51.68 \pm 30.84$, among the top 40 prompts. Some structures, like individual femur pones and gluteous muscle groups, are more easily localized, with a centroid error of less than 40\,mm, while others, like the liver, pose more of a challenge, possibly due to their asymmetric shape. Table~\ref{tab:twin-results} in the supplement details these results. 

On the Loop-X, 3D collimation is achieved by defining a bounding box in the patient coordinate system which all future acquisitions should collimate to. We evaluate our digital twin's value for collimation in terms of the bounding box statistics in the patient coordinate system of the post-study CT. The bounding box recall provides a measure of how much of the desired anatomy is not present in the collimation volume, while the precision measures how tightly our digital twin is able to isolate the desired anatomy. Over the tested prompts, the average precision was $0.26 \pm 0.22$ and the average recall was $0.70 \pm 0.26$, indicating high coverage of the desired structure and reasonable isolation even when cropped tightly. Fig.~\ref{fig:likelihood-results} shows the performance for each prompt tested.

\begin{figure}[t]
  \centering
  \includegraphics[width=\linewidth]{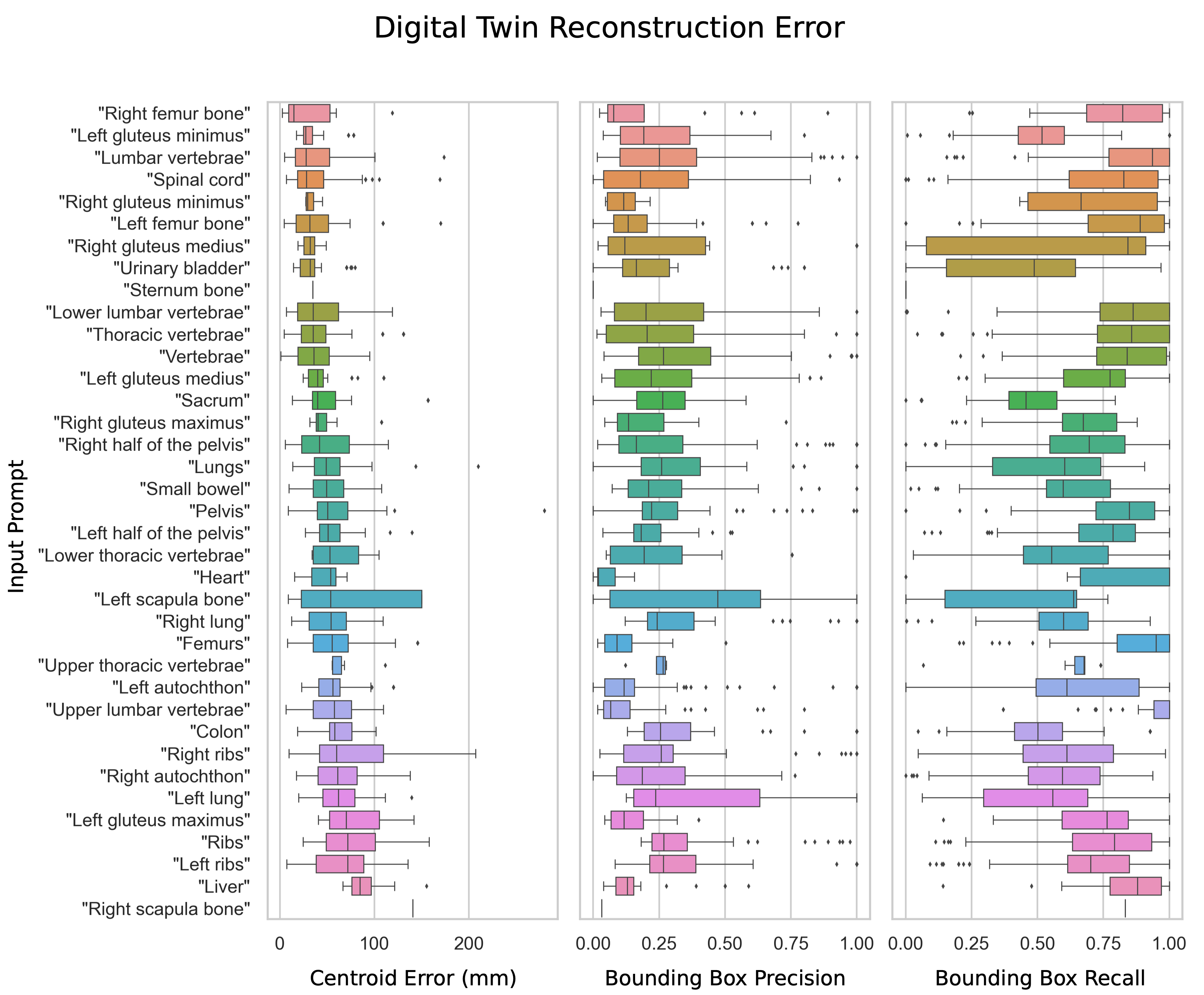}
  \caption{The reconstruction error of the digital twin in terms of localization and collimation of desired structures. We observe a tendency toward better localization and isolation for structures which FluoroSAM segments more easily.}
  \label{fig:likelihood-results}
\end{figure}

\subsection{Real-time Cadaver Study}
\label{sec:cadaver-study}

In a real-time study, our approach was able to visualize, localize, and collimate to anatomical structures based solely on voice control with a high success rate. Success was evaluated based on the end-to-end success, including failures due to poor transcription or movement constraints. Over 158 prompts, 38 / 46 (82.6\%) visualization actions, 17 / 28 (60.7\%) collimations, and 10 / 12 (83.33\%) were successful. Other low-level actions, like ``Take a shot'' or ``Roll over 30 degrees'' accounted for the remainder of the prompts. Of these, 40 were spoken by a person with an Indian accent and 118 by a person with an American accent, with similar success rates of 87.5\% and 82.2\%, respectively. We observed some accent-specific transcription errors of medical words, like ``Collimate'' as ``column 8'' when spoken in isolation. Fig.~\ref{fig:overview}a and the video supplement show the study, which had an overall success rate of 83.54\%.

\section{Discussion and Conclusion}
\label{sec:discussion}

% \cite{killeen2024stand} for large scale human subject trials of VUIs in simulation

% Eventually, simulation of X-ray guided workflows like in \cite{killeen2023pelphix} could be used to anticipate and actively prepare for surgeon commands. In the case of \cite{killeen2023pelphix}, that preparation is procedure-specific, but the incorporation of foundation models as in here could allow for more general surgeon-AI collaboration.

% Peronsalization may also be a possibility using LLMs~\cite{wu2023tidybot}.

% Limtations in practice. English based language model. Human comouter interaction studies needed, Standing on ones shoes paper~\cite{killeen2024stand}. 
% here we showed its possible to have a general system
% Nobody has collected data on the full broad on commanding C-arm 
% \cite{demir2022pocap} a dataset in the domain of interventional radiology workflow for the development of a smart OR speech assistant able to identify surgeons' command words during operations. It consists of X-ray images, RGB ambient videos, speech uttered by the surgeons, and system commands given by medical personnel to operate devices in the OR.
% "We claim that a corpus for surgical workflow analysis can benefit from multimodal data, including not only images and videos, but also speech uttered by the surgeons in their language".

There are notable limitations in the approach outlined here. As we observed in our study, the performance of the foundation model, FluoroSAM, is severely limited when used without additional point prompts. Although we identified a number of anatomical structures in the torso region which it was able to localize effectively, a more capable model would lend itself to a wider variety of tasks and anatomies. Additionally, speech-to-text systems exhibit notable bias toward users with different accents and languages, as we observed in our study, and make transcription errors when lacking the medical context. Mechanical solutions, like a foot-pedal mute switch, may reduce the need to differentiate intentional commands from other communications among the surgical team, but it is an inelegant solution. As LLMs become faster and more capable, they are exhibiting a greater understanding of users' intent in a more general context, including when the system is being addressed. %\cite{robla2017working}
Moving forward, there is a need for more detailed evaluation of potential human-robot interactions in the operating room. This is to better understand physicians' needs and how they may express them throughout a given procedure. Initial efforts have collected data for surgical workflow analysis that includes system commands by medical personnel to operate devices in the OR~\cite{demir2022pocap}, but patient privacy concerns prevent open data sharing. Additionally, data from current ORs is fundamentally limited because it contains surgeon-technologist interactions, but validation is needed before intelligent systems can be deployed in patient care. One avenue, which promises to enhance both data collection and human-centered design, is the use of virtual reality environments for image-guided procedures~\cite{killeen2024stand}. These allow surgeons and technologists to work through communication challenges with intelligent assistance systems in a more accessible, scalable way.

We have shown that a language-aligned foundation model can be used to facilitate flexible control of a robotic C-arm device. As images are acquired in the course of a procedure, our approach continually updates a digital twin to allow visualization, collimation, and viewfinding for desired anatomies based on natural language prompting. In a cadaver study, the fully integrated system successfully interprets commands and performs requested actions.

\backmatter

% \bmhead{Supplementary information}

% The video supplement includes video from our real-time cadaver study.

% \bmhead{Acknowledgements}

% \section*{Declarations}

% Some journals require declarations to be submitted in a standardised format. Please check the Instructions for Authors of the journal to which you are submitting to see if you need to complete this section. If yes, your manuscript must contain the following sections under the heading `Declarations':

% \begin{itemize}
% \item Conflict of interest/Competing interests (check journal-specific guidelines for which heading to use)
% \item Ethics approval and consent to participate
% \item Consent for publication
% \item Materials availability
% \end{itemize}

\bibliography{references}
% \bibliography{ref-short}

\begin{appendices}

\newpage

\section{Additional Results}\label{sec:extra-results}

\begin{table}[h]
  \centering
  \begin{tabular}{lrr}
    \toprule
    Prompt & DICE & Centroid Error (mm) \\
    \midrule
    ``Vertebrae'' & 0.78 $\pm$ 0.15 & 21.80 $\pm$ 24.04 \\
    ``Lumbar vertebrae'' & 0.72 $\pm$ 0.18 & 29.12 $\pm$ 35.00 \\
    ``Lower lumbar vertebrae'' & 0.68 $\pm$ 0.19 & 37.57 $\pm$ 46.94 \\
    ``Right lung'' & 0.67 $\pm$ 0.22 & 42.18 $\pm$ 35.16 \\
    ``Femurs'' & 0.66 $\pm$ 0.17 & 43.45 $\pm$ 29.83 \\
    ``Thoracic vertebrae'' & 0.64 $\pm$ 0.20 & 51.16 $\pm$ 43.23 \\
    ``Left femur bone'' & 0.64 $\pm$ 0.26 & 41.39 $\pm$ 57.24 \\
    ``Lungs'' & 0.64 $\pm$ 0.16 & 45.96 $\pm$ 28.75 \\
    ``Spinal cord'' & 0.64 $\pm$ 0.24 & 35.47 $\pm$ 32.48 \\
    ``Left lung'' & 0.57 $\pm$ 0.22 & 41.86 $\pm$ 17.36 \\
    ``Ribs'' & 0.54 $\pm$ 0.12 & 50.62 $\pm$ 25.30 \\
    ``Pelvis'' & 0.53 $\pm$ 0.16 & 67.35 $\pm$ 41.86 \\
    ``Left ribs'' & 0.53 $\pm$ 0.12 & 44.15 $\pm$ 29.74 \\
    ``Left scapula bone'' & 0.53 $\pm$ 0.24 & 38.03 $\pm$ 40.84 \\
    ``Right femur bone'' & 0.51 $\pm$ 0.33 & 66.77 $\pm$ 77.62 \\
    ``Left half of the pelvis'' & 0.51 $\pm$ 0.13 & 69.75 $\pm$ 42.68 \\
    ``Right half of the pelvis'' & 0.50 $\pm$ 0.21 & 66.92 $\pm$ 46.42 \\
    ``Right scapula bone'' & 0.49 $\pm$ 0.15 & 40.72 $\pm$ 22.21 \\
    ``Right ribs'' & 0.48 $\pm$ 0.21 & 54.42 $\pm$ 50.28 \\
    ``Left autochthon'' & 0.47 $\pm$ 0.24 & 52.93 $\pm$ 40.06 \\
    ``Small bowel'' & 0.42 $\pm$ 0.20 & 71.95 $\pm$ 48.55 \\
    ``Right autochthon'' & 0.42 $\pm$ 0.25 & 54.40 $\pm$ 36.07 \\
    ``Upper lumbar vertebrae'' & 0.42 $\pm$ 0.18 & 77.13 $\pm$ 35.25 \\
    ``Right gluteus maximus'' & 0.40 $\pm$ 0.23 & 73.43 $\pm$ 78.34 \\
    ``Left gluteus minimus'' & 0.38 $\pm$ 0.23 & 45.63 $\pm$ 44.00 \\
    ``Left gluteus maximus'' & 0.35 $\pm$ 0.21 & 72.05 $\pm$ 36.12 \\
    ``Sacrum'' & 0.35 $\pm$ 0.19 & 62.85 $\pm$ 63.14 \\
    ``Upper thoracic vertebrae'' & 0.35 $\pm$ 0.22 & 141.87 $\pm$ 66.56 \\
    ``Urinary bladder'' & 0.34 $\pm$ 0.16 & 37.76 $\pm$ 30.95 \\
    ``Heart'' & 0.33 $\pm$ 0.19 & 58.55 $\pm$ 24.90 \\
    ``Left gluteus medius'' & 0.32 $\pm$ 0.19 & 72.97 $\pm$ 61.26 \\
    ``Lower thoracic vertebrae'' & 0.27 $\pm$ 0.19 & 159.99 $\pm$ 63.95 \\
    ``L5 vertebra bone'' & 0.25 $\pm$ 0.25 & 50.71 $\pm$ 48.80 \\
    ``Right clavicle bone'' & 0.25 $\pm$ 0.20 & 46.99 $\pm$ 59.10 \\
    ``L3 vertebra bone'' & 0.21 $\pm$ 0.21 & 53.86 $\pm$ 41.80 \\
    ``Right kidney'' & 0.19 $\pm$ 0.09 & 51.06 $\pm$ 23.15 \\
    ``Right gluteus minimus'' & 0.19 $\pm$ 0.19 & 62.42 $\pm$ 41.89 \\
    ``Left clavicle bone'' & 0.17 $\pm$ 0.18 & 33.27 $\pm$ 24.40 \\
    ``Sternum bone'' & 0.17 $\pm$ 0.20 & 70.62 $\pm$ 38.32 \\
    ``L4 vertebra bone'' & 0.16 $\pm$ 0.15 & 49.40 $\pm$ 38.92 \\
    ``Kidneys'' & 0.16 $\pm$ 0.09 & 83.09 $\pm$ 30.16 \\
    Avg. & 0.47 $\pm$ 0.25 & 58.87 $\pm$ 50.29 \\
  \bottomrule
\end{tabular}
  \caption{FluoroSAM Single-image Performance}
  \label{tab:fluorosam-results}
\end{table}

\begin{table}[h]
    \centering
    \begin{tabular}{lrrr}
    \toprule
    Prompt & 3D Centroid Error (mm) & B. Box Precision & B. Box Recall \\
    \midrule
    ``Right femur bone'' & 27.50 $\pm$ 28.07 & 0.17 $\pm$ 0.22 & 0.77 $\pm$ 0.23 \\
    ``Right gluteus medius'' & 31.83 $\pm$ 8.97 & 0.33 $\pm$ 0.39 & 0.58 $\pm$ 0.41 \\
    ``Left gluteus minimus'' & 32.44 $\pm$ 15.87 & 0.24 $\pm$ 0.18 & 0.51 $\pm$ 0.20 \\
    ``Right gluteus minimus'' & 32.60 $\pm$ 7.19 & 0.12 $\pm$ 0.07 & 0.70 $\pm$ 0.27 \\
    % ``Sternum bone'' & 34.63 $\pm$ nan & 0.00 $\pm$ nan & 0.00 $\pm$ nan \\
    ``Lumbar vertebrae'' & 35.28 $\pm$ 28.23 & 0.30 $\pm$ 0.25 & 0.84 $\pm$ 0.20 \\
    ``Urinary bladder'' & 35.29 $\pm$ 19.17 & 0.25 $\pm$ 0.24 & 0.44 $\pm$ 0.31 \\
    ``Spinal cord'' & 36.35 $\pm$ 26.84 & 0.24 $\pm$ 0.22 & 0.74 $\pm$ 0.29 \\
    ``Vertebrae'' & 37.77 $\pm$ 22.04 & 0.32 $\pm$ 0.22 & 0.82 $\pm$ 0.18 \\
    ``Left femur bone'' & 37.97 $\pm$ 30.08 & 0.18 $\pm$ 0.17 & 0.80 $\pm$ 0.24 \\
    ``Lower lumbar vertebrae'' & 40.27 $\pm$ 26.86 & 0.26 $\pm$ 0.23 & 0.81 $\pm$ 0.24 \\
    ``Thoracic vertebrae'' & 43.25 $\pm$ 30.24 & 0.25 $\pm$ 0.25 & 0.80 $\pm$ 0.25 \\
    ``Heart'' & 45.40 $\pm$ 19.16 & 0.05 $\pm$ 0.06 & 0.81 $\pm$ 0.30 \\
    ``Left gluteus medius'' & 47.11 $\pm$ 24.19 & 0.28 $\pm$ 0.26 & 0.68 $\pm$ 0.22 \\
    ``Right gluteus maximus'' & 49.00 $\pm$ 20.64 & 0.20 $\pm$ 0.16 & 0.62 $\pm$ 0.23 \\
    ``Sacrum'' & 49.19 $\pm$ 31.53 & 0.25 $\pm$ 0.14 & 0.46 $\pm$ 0.17 \\
    ``Right half of the pelvis'' & 49.53 $\pm$ 29.42 & 0.29 $\pm$ 0.30 & 0.65 $\pm$ 0.25 \\
    ``Right lung'' & 51.20 $\pm$ 25.85 & 0.34 $\pm$ 0.23 & 0.58 $\pm$ 0.22 \\
    ``Small bowel'' & 53.43 $\pm$ 24.87 & 0.26 $\pm$ 0.19 & 0.60 $\pm$ 0.22 \\
    ``Upper lumbar vertebrae'' & 53.67 $\pm$ 28.25 & 0.14 $\pm$ 0.18 & 0.93 $\pm$ 0.15 \\
    ``Left half of the pelvis'' & 53.98 $\pm$ 18.94 & 0.21 $\pm$ 0.09 & 0.73 $\pm$ 0.21 \\
    ``Left autochthon'' & 54.92 $\pm$ 19.04 & 0.17 $\pm$ 0.22 & 0.64 $\pm$ 0.27 \\
    ``Lungs'' & 55.24 $\pm$ 34.91 & 0.35 $\pm$ 0.28 & 0.54 $\pm$ 0.29 \\
    ``Femurs'' & 58.49 $\pm$ 33.84 & 0.10 $\pm$ 0.08 & 0.87 $\pm$ 0.19 \\
    ``Pelvis'' & 58.82 $\pm$ 34.77 & 0.29 $\pm$ 0.19 & 0.79 $\pm$ 0.21 \\
    ``Lower thoracic vertebrae'' & 61.17 $\pm$ 26.53 & 0.23 $\pm$ 0.21 & 0.57 $\pm$ 0.26 \\
    ``Colon'' & 62.85 $\pm$ 18.92 & 0.33 $\pm$ 0.22 & 0.50 $\pm$ 0.18 \\
    ``Upper thoracic vertebrae'' & 64.14 $\pm$ 16.54 & 0.25 $\pm$ 0.04 & 0.62 $\pm$ 0.19 \\
    ``Right autochthon'' & 64.55 $\pm$ 28.27 & 0.23 $\pm$ 0.17 & 0.57 $\pm$ 0.24 \\
    ``Left lung'' & 64.90 $\pm$ 28.98 & 0.36 $\pm$ 0.26 & 0.52 $\pm$ 0.32 \\
    ``Left ribs'' & 69.80 $\pm$ 33.14 & 0.35 $\pm$ 0.26 & 0.68 $\pm$ 0.24 \\
    ``Left scapula bone'' & 70.56 $\pm$ 58.55 & 0.41 $\pm$ 0.33 & 0.48 $\pm$ 0.29 \\
    ``Right ribs'' & 71.77 $\pm$ 44.00 & 0.32 $\pm$ 0.29 & 0.59 $\pm$ 0.25 \\
    ``Ribs'' & 76.16 $\pm$ 34.18 & 0.35 $\pm$ 0.21 & 0.73 $\pm$ 0.24 \\
    ``Left gluteus maximus'' & 78.10 $\pm$ 29.58 & 0.14 $\pm$ 0.08 & 0.73 $\pm$ 0.17 \\
    ``Liver'' & 89.05 $\pm$ 18.40 & 0.16 $\pm$ 0.14 & 0.85 $\pm$ 0.18 \\
    % ``Right scapula bone'' & 140.70 $\pm$ nan & 0.03 $\pm$ nan & 0.83 $\pm$ nan \\
    Avg. & 51.68 $\pm$ 30.84 & 0.26 $\pm$ 0.22 & 0.70 $\pm$ 0.26 \\
    \bottomrule
    \end{tabular}
    \caption{3D Digital Twin Reconstruction for Selected Prompts}
    \label{tab:twin-results}
\end{table}




\end{appendices}


\end{document}